%% file: main.tex
\documentclass{article} %

\input{math_commands.tex}

\usepackage[dvipsnames]{xcolor}
\usepackage{hyperref}
\hypersetup{
    colorlinks=true,
	citecolor=MidnightBlue,
	linkcolor=OrangeRed,
	urlcolor=OrangeRed
}
\usepackage{iclr2026_conference,times}
\usepackage[utf8]{inputenc}
\usepackage[T1]{fontenc}
\usepackage{url}
\usepackage{booktabs}
\usepackage{amsfonts}
\usepackage{amsmath}
\usepackage{nicefrac}
\usepackage{microtype}
\usepackage{graphicx}
\usepackage{multirow}
\usepackage{natbib}
\usepackage{amssymb}
\usepackage{makecell}
\usepackage{cite}
\usepackage{todonotes}
\usepackage{tabularx}
\usepackage{subcaption}
\usepackage{wrapfig}
\usepackage{marvosym}
\usepackage{enumitem}

\usepackage[capitalise,nameinlink]{cleveref}
\crefname{figure}{Fig.}{Figs.}
\Crefname{figure}{Figure}{Figures}
\crefname{table}{Tab.}{Tabs.}
\Crefname{table}{Table}{Tables}

\title{IVGT: Implicit Visual Geometry Transformer for Neural Scene Representation}

\author{Yuqi Wu$^{\!}$\thanks{Equal contributions.  
  \textsuperscript{\textdagger}Project leader.
  } \quad
  Tianyu Hu$^{*}$ \quad
  Wenzhao Zheng$^{*,\dagger}$ \quad \\
  \textbf{Yuanhui Huang} \quad
  \textbf{Haowen Sun} \quad
  \textbf{Jie Zhou} \quad
  \textbf{Jiwen Lu}
  \vspace{2mm}
  \\
  Intelligent Vision Group, Tsinghua University
  \vspace{1mm}
  \\
\textbf{Page}: \url{https://wzzheng.net/IVGT/}\\
\textbf{Code}: \url{https://github.com/wzzheng/IVGT/}
}

\iclrfinalcopy
\begin{document}

\maketitle

\input{chapters/0_abstract.tex}
\input{chapters/1_introduction}
\input{chapters/2_related}

\input{chapters/3_method}
\input{chapters/4_exp}
\input{chapters/5_conclusion}

\newpage

\input{main.bbl}
\clearpage

\end{document}

%% file: math_commands.tex
\usepackage{amsmath,amsfonts,bm}

\def\eqref#1{equation~\ref{#1}}

\def\1{\bm{1}}

\DeclareMathAlphabet{\mathsfit}{\encodingdefault}{\sfdefault}{m}{sl}
\SetMathAlphabet{\mathsfit}{bold}{\encodingdefault}{\sfdefault}{bx}{n}

%% file: chapters/0_abstract.tex
\vspace{-8mm}
\begin{center}
     \centering
     \includegraphics[width=1\linewidth]{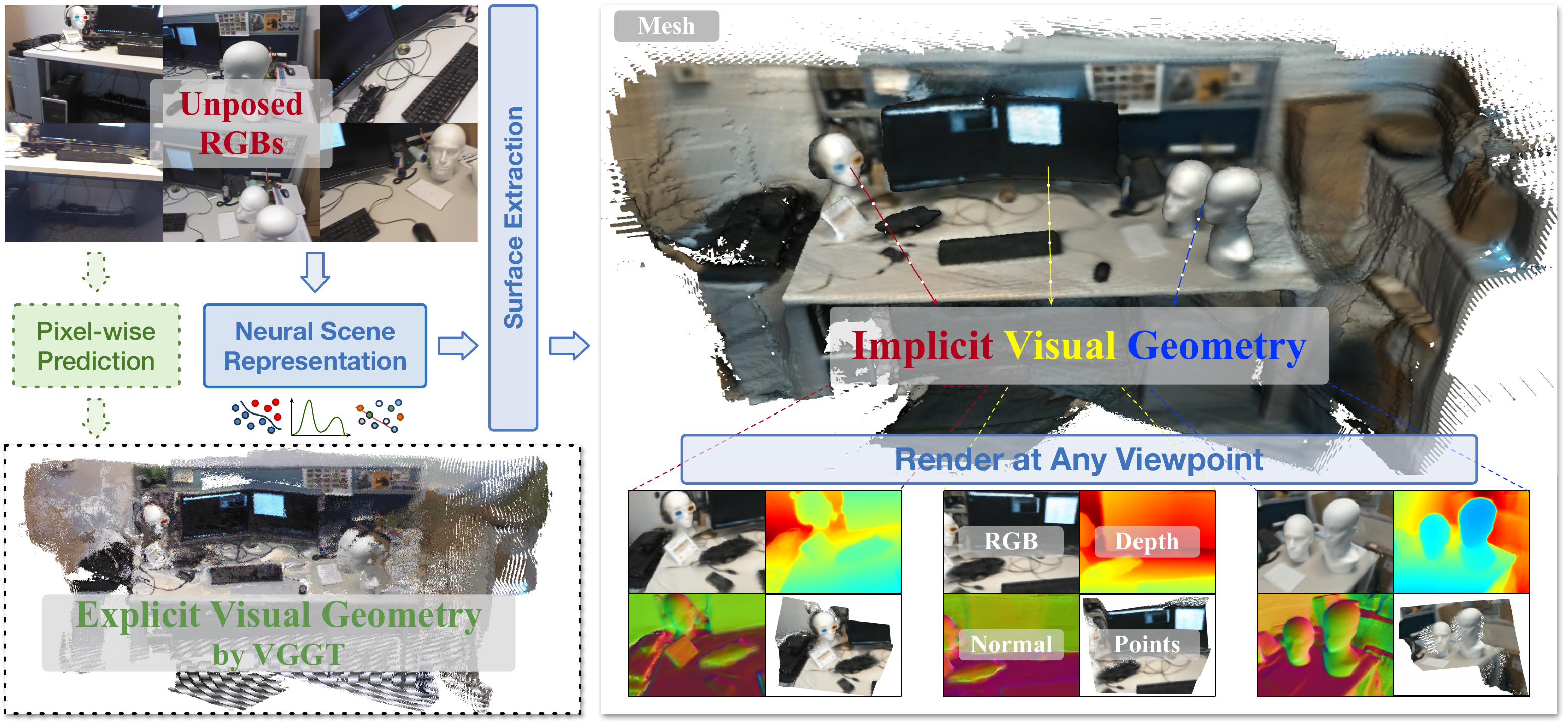}
     \vspace{-7mm}
     \captionof{figure}{    
    \textbf{IVGT} implicitly models coherent 3D geometry and appearance from pose-free multi-view images in one feedforward pass.
      It learns a continuous neural scene representation in a global canonical coordinate system, enabling various downstream tasks including mesh reconstruction, novel-view synthesis, and surface estimation across diverse scenes.
 }
 \label{fig:teaser}
 \end{center}
\vspace{-1mm}

\begin{abstract}
\vspace{-2mm}
Reconstructing coherent 3D geometry and appearance from unposed multi-view images is a fundamental yet challenging problem in computer vision.
Most existing visual geometry foundation models predict explicit geometry by regressing pixel-aligned pointmaps, often suffering from redundancy and limited geometric continuity.
We propose \textbf{IVGT}, an \textbf{I}mplicit \textbf{V}isual \textbf{G}eometry \textbf{T}ransformer that implicitly models continuous and coherent geometry from pose-free multi-view images.
This formulation learns a continuous neural scene representation in a canonical coordinate system and supports continuous spatial queries at any 3D positions, retrieving local features to predict signed distance (SDF) values and colors using lightweight decoders.
It allows direct extraction of continuous and coherent surface geometry, enabling rendering of RGB images, depth maps, and surface normal maps from arbitrary viewpoints.
We train IVGT via multi-dataset joint optimization with 2D supervision and 3D geometric regularization.
IVGT demonstrates generalization across scenes and achieves strong performance on various tasks, including mesh and point cloud reconstruction, novel view synthesis, depth and surface normal estimation, and camera pose estimation.
\end{abstract}

\vspace{-5mm}

%% file: chapters/1_introduction.tex
\section{Introduction}

Reconstructing 3D scenes from multi-view images~\citep{snavely2006photo, furukawa2009accurate, schoenberger2016sfm} has long been a central problem in computer vision, with wide applications in robotics~\citep{wang2024embodiedscan,embodiedocc} and autonomous driving~\citep{huang2023tri, dvgt, dvgt2}.
Given multi-view images, the goal is to recover a representation that captures both accurate geometry and view-consistent appearance, while supporting flexible rendering and downstream reasoning~\citep{mildenhall2021nerf, yu2021pixelnerf}.
However, learning such a unified 3D representation from multi-view images remains challenging, as the reconstructed geometry should be consistent across viewpoints and exhibit continuous and coherent surface structures.

Conventional methods estimate sparse geometry and camera poses via structure-from-motion (SfM)~\citep{snavely2006photo, frahm2010building, schoenberger2016sfm}, followed by dense reconstruction using multi-view stereo (MVS)~\citep{furukawa2009accurate, gu2020cascade}, neural radiance fields (NeRF)~\citep{mildenhall2021nerf, barron2022mip, muller2022instant}, or 3D Gaussian splatting (3DGS)~\citep{kerbl3Dgaussians}.
While some methods~\citep{yariv2021volume, wang2021neus, yu2022monosdf,neuralrecon} have explored modeling scenes as an implicit signed distance field (SDF) for extraction of continuous and smooth surfaces, typically require long per-scene optimization or accurate input camera poses.
The recent emergence of visual geometry foundation models~\citep{wang2024dust3r,wang2025vggtvisualgeometrygrounded,wu2025point3r,zhuo2025streaming} has significantly revolutionized the paradigm of 3D reconstruction.
By adopting an end-to-end formulation and large-scale pretraining, they directly regress multi-view observations into a shared coordinate system without requiring explicit camera poses~\citep{wang20243d, cut3r, wu2025point3r, zhuo2025streaming, chen2025ttt3r, lan2025stream3r, fang2025dens3r} .
They decode pixel-aligned features into explicit 3D point coordinates, simplifying the previously complex reconstruction pipeline.
However, this explicit pixel-aligned formulation often leads to redundant representations and lacks geometric continuity.
Furthermore, rendering novel views from this representation requires additional modeling~\citep{liu2025worldmirror} or post-processing.
As a result, learning an visual geometry foundation model that supports coherent surface reconstruction and continuous rendering is of great value but remains unexplored.

To bridge this gap, we propose an \textbf{I}mplicit \textbf{V}isual \textbf{G}eometry \textbf{T}ransformer (\textbf{IVGT}) that learns a global neural scene representation from pose-free multi-view images to implicitly model continuous and coherent geometry, as shown in \Cref{fig:teaser}.
Specifically, a transformer backbone jointly processes tokens from all input views through alternating intra-view and cross-view attention, aggregating multi-view observations into a unified canonical representation without requiring explicit camera poses as input.
This representation implicitly aligns multiple views within a shared coordinate system and supports continuous spatial queries over the 3D space.
Given an arbitrary 3D query point, the model projects it onto each input view to retrieve the corresponding pixel-aligned features, which are further augmented with view-relative depth encodings to disambiguate points lying along the same projection ray.
The aggregated local spatial feature is then decoded by cascaded MLP decoders: a geometry decoder predicts the signed distance function (SDF) value at the query point, and an appearance decoder takes the SDF gradient (i.e., the surface normal) and the viewing direction to predict view-dependent color.
This implicit SDF-based formulation enables unified modeling of geometry and appearance under a single continuous representation.
The predicted SDF values are converted to densities to support differentiable volume rendering, allowing the model to render RGB images, depth maps, and surface normal maps from arbitrary novel viewpoints.
Moreover, IVGT achieves efficient and high-quality surface extraction via the Marching Cubes algorithm~\citep{lorensen1998marching}, directly yielding continuous mesh representations of the scene, which is inherently difficult to achieve with discrete and pixel-aligned explicit outputs.

To fully leverage large-scale datasets with only 2D observations and pose annotations, we design a multi-stage training strategy that relies solely on 2D supervision and 3D geometric regularization.
This training process progressively enables the model to learn view-consistent rendering and stable surface reconstruction.
Through joint training on object-level and scene-level datasets, our model generalizes across diverse scenes and performs inference in a single forward pass, without any test-time optimization or post-processing.
We evaluate our method on diverse tasks including mesh and pointmap reconstruction, novel view synthesis, depth and surface normal estimation.
Extensive experiments demonstrate that our model achieves strong performance in consistent and continuous geometry reconstruction from pose-free multi-view inputs.

%% file: chapters/2_related.tex
\section{Related Work}

\textbf{Multi-view 3D Reconstruction.}
Reconstructing 3D geometry from multi-view images is a fundamental problem in computer vision. 
Conventional pipelines typically decompose this problem into multiple stages, where sparse geometry and camera poses are first estimated via structure-from-motion (SfM)~\citep{agarwal2011building, crandall2012sfm, wu2013towards, wilson2014robust, schoenberger2016sfm}, followed by dense reconstruction using multi-view stereo (MVS)~\citep{furukawa2015multi, colmapmvs, Wei_2021_ICCV, fu2022geo}, neural radiance fields (NeRF)~\citep{mildenhall2021nerf, barron2022mip, chen2022tensorf}, or 3D Gaussian splatting (3DGS)~\citep{kerbl3Dgaussians}. 
While these methods can produce high-quality results, they rely heavily on accurate camera pose estimation and involve complex cascaded stages, which hinder the end-to-end optimization and result in limited robustness in challenging scenarios.
Recent learning-based methods~\citep{wang2024dust3r, cut3r, wu2025point3r, zhuo2025streaming, wang2025vggtvisualgeometrygrounded} aim to simplify this process by directly learning geometry from images. 
They remove the dependency on camera poses by jointly reasoning about multi-view correspondences, enabling end-to-end training and large-scale generalization. 
However, they primarily focus on predicting geometry in view-aligned explicit forms such as depth maps or point maps, and do not directly model continuous scene representations that support flexible querying and rendering. 
This limits their ability to unify multi-view understanding with downstream tasks such as novel view synthesis and surface reconstruction.

\textbf{Neural Implicit Scene Representations.}
Neural implicit representations model 3D scenes as continuous functions parameterized by neural networks. 
Early methods~\citep{xie2022neural, chen2019learning, mescheder2019occupancy} rely on an MLP to represent geometry and appearance of a single scene, achieving strong reconstruction quality at the object level but struggling to scale to complex scenes due to limited model capacity and slow optimization. 
Subsequent works~\citep{chibane2020implicit, martel2021acorn, takikawa2021neural, yu2022monosdf} introduce hybrid designs that combine MLPs with multi-resolution voxel grids to capture finer geometric details more efficiently. 
While these methods improve representation capacity, their computational costs grow significantly with scene size.
More importantly, most neural implicit approaches are optimized on a per-scene basis, requiring long training times and preventing direct generalization across different scenarios. 
As a result, although they provide continuous and flexible representations, their applicability in large-scale or real-time settings remains limited, motivating the need for representations that can be learned across scenes while retaining the advantages of continuous geometry modeling.

\textbf{Generalizable Neural Fields.}
The transformer architecture has demonstrated strong generalizable capability in aggregating multi-scale contextual information across a wide range of dense prediction tasks, including segmentation~\citep{lavt}, depth estimation~\citep{ranftl21dpt}, and optical flow~\citep{teed2020raft}.
Motivated by this success, recent works~\citep{yu2021pixelnerf, wang2021ibrnet, chen2021mvsnerf, neuralrecon, yu2026infinidepth} explore generalizable neural fields that can be trained across multiple scenes and adapted to new scenes with minimal finetuning, conditioning 3D predictions on transformer-extracted image features. 
These methods enable fast inference and improved generalization across scenes. 
However, they typically rely on known camera poses to establish geometric relationships across views~\citep{neuralrecon, guo2022neural, sayed2022simplerecon}. 
In addition, since these methods are primarily designed for novel view synthesis, they focus more on appearance modeling than on enforcing consistent multi-view geometry~\citep{jeong2026view}, which can limit the quality of reconstructed surfaces. 
Differently, we train an implicit visual geometry transformer that aggregates a neural scene representation from pose-free images, enabling not only continuous spatial queries to render RGB images, depth maps, and surface normal maps from arbitrary viewpoints, but also to extract a continuous surface representation.

%% file: chapters/3_method.tex
\section{Proposed Approach}

We introduce IVGT, an implicit visual geometry transformer that learns a neural scene representation within a canonical coordinate system from pose-free multi-view images.
We first introduce the problem in Sec.~\ref{3.1}, then elaborate on our pose-free neural scene representation in Sec.~\ref{3.2}, followed by rendering and surface extraction in Sec.~\ref{3.3}, and finally describe the training setup in Sec.~\ref{3.4}.

\begin{figure}[t]  
    \centering  
    \includegraphics[width=\textwidth]{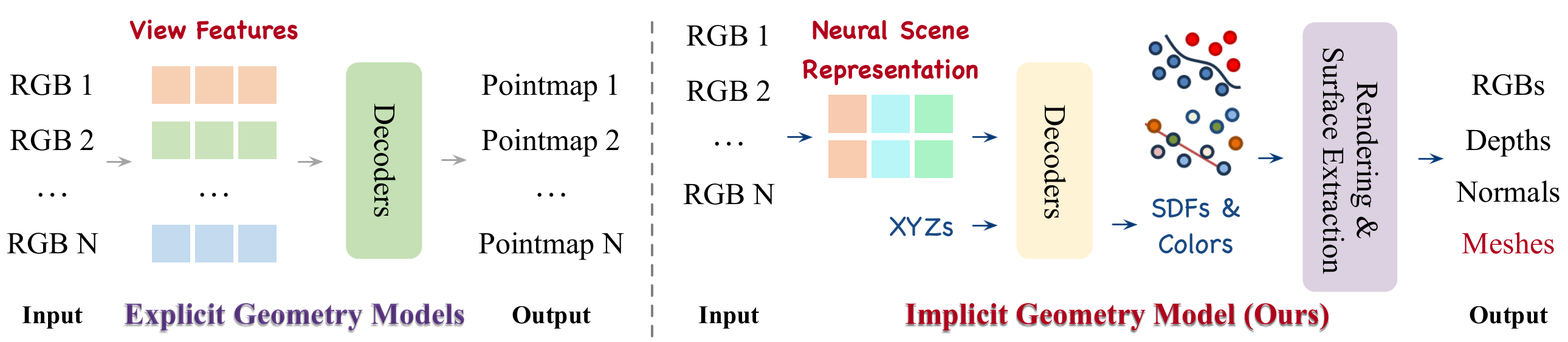}
    \vspace{-7mm}
    \caption{\textbf{Explicit vs.\ implicit visual geometry paradigms.} Existing explicit models decode per-pixel 3D point coordinates for each input view independently, producing discrete and view-indexed pointmaps. Our implicit model instead learns a continuous 3D field from which any spatial location can be queried, enabling direct surface extraction and novel view rendering without post-processing.}
    \label{fig:methodcompare} 
    \vspace{-5mm}
\end{figure}

\subsection{Implicit Visual Geometry Modeling}
\label{3.1}

Visual geometry foundation models aim to recover the 3D structure of a scene from unposed multi-view images in a single forward pass, without per-scene optimization or known camera parameters.
By learning from large-scale multi-view data, such models acquire generalizable geometric priors and enable fast, robust reconstruction across diverse scenes and domains~\citep{wang2024dust3r, wang2025vggtvisualgeometrygrounded, wu2025point3r, zhuo2025streaming}.
A central design choice in these models is how scene geometry is \emph{represented}: the representation determines what geometric queries are supported, whether the output is continuous or discrete, and how rendering and surface extraction can be performed.

\textbf{Explicit Visual Geometry.}
As illustrated in \Cref{fig:methodcompare}, existing visual geometry foundation models~\citep{wang2024dust3r,wang2025vggtvisualgeometrygrounded,wu2025point3r} follow an \textbf{explicit} paradigm, where the model directly decodes a pixel-aligned pointmap for each input view.
Formally, given $N$ input images $\{I_i\}_{i=1}^N$, these methods produce a pointmap $\mathcal{P}_i \in \mathbb{R}^{H \times W \times 3}$ per view:
\begin{equation}
    \mathcal{P}_i = h_i(\mathcal{F}), \quad \mathcal{F} = \mathrm{Enc}(\{I_i\}_{i=1}^N),
\end{equation}
where $h_i$ is a per-view decoder and $\mathcal{F}$ is the encoded multi-view feature.
The full scene geometry is thus a finite collection of $N \times H \times W$ discrete 3D points, each anchored to a specific pixel in a specific view.
While effective for coarse geometry estimation and camera pose prediction, this explicit formulation has two fundamental limitations.
\textbf{First}, the representation is inherently \emph{discrete}: geometry is only defined at pixel-aligned locations, so arbitrary 3D points cannot be queried and no continuous surface is implied.
\textbf{Second}, the same physical surface point $\mathbf{x}$ may appear in multiple views and be predicted redundantly, yielding potentially inconsistent estimates across views.
As a result, extracting a continuous surface or rendering from a novel viewpoint both require additional modeling or non-trivial post-processing.

\textbf{Implicit Visual Geometry (Ours).}
In contrast, we propose to learn a continuous 3D field from pose-free multi-view images.
Rather than decoding a fixed set of pixel-aligned points, the model learns a global neural scene representation $\mathcal{F}$ that can be queried at \emph{any} 3D location $\mathbf{x} \in \mathbb{R}^3$:
\begin{equation}
    \bigl(\hat{s}(\mathbf{x}),\, \hat{\mathbf{c}}(\mathbf{x}, \mathbf{v})\bigr)
    = \mathrm{Dec}\!\left(\mathbf{x},\, \mathbf{v};\, \mathcal{F}\right),
    \label{eq:implicit_query}
\end{equation}
where $\hat{s}(\mathbf{x}) \in \mathbb{R}$ is the signed distance value and $\hat{\mathbf{c}}(\mathbf{x}, \mathbf{v}) \in \mathbb{R}^3$ is the view-dependent color for viewing direction $\mathbf{v}$.
The scene surface is uniquely defined as the zero level set $\mathcal{S} = \{\mathbf{x} \in \mathbb{R}^3 : \hat{s}(\mathbf{x}) = 0\}$, which is continuous and view-count-independent.
Volume rendering along any ray directly yields RGB, depth, and surface normal, and the iso-surface can be extracted via Marching Cubes~\citep{lorensen1998marching} without post-processing.
The full input--output mapping of IVGT is:
\begin{equation}
     \bigl(\mathcal{F},\, (D_{i},\, \mathbf{g}_{i})_{i=1}^{N}\bigr) = f\!\left(\{I_i\}_{i=1}^{N}\right),
\end{equation}
where $\mathbf{g}_{i} \in \mathbb{R}^{9}$ and $D_{i} \in \mathbb{R}^{H\times W}$ are auxiliary camera parameters and depth maps predicted alongside $\mathcal{F}$.
We take the first image as the reference frame, and the model is permutation equivariant for all remaining views.

\subsection{Pose-free Continuous Neural Scene Representation}
\label{3.2}

\textbf{Image Encoder and Global Feature Extraction.}
We adopt a transformer-based backbone to extract multi-view image features and learn a global scene representation.
Given a set of input images \((\mathit{I}_{i})_{i=1}^{N}\), each image is first converted into a set of tokens by DINO~\citep{oquab2023dinov2}.
All tokens from different views are then jointly processed by a multi-layer transformer that alternates between frame-wise and global self-attention~\citep{wang2025vggtvisualgeometrygrounded}, enabling both intra-view feature learning and cross-view information aggregation.
This design allows the model to implicitly align multi-view observations without requiring explicit camera poses.
In addition to feature extraction, the backbone also predicts auxiliary geometric quantities for each view, including depth map \(\mathit{D}_{i}\) and camera parameters \(\mathbf{g}_{i}\), which provide useful geometric cues for downstream representation learning.
We obtain a global scene representation \(\mathcal{F}=\{\mathit{F}_i\}_{i=1}^{N}\) from this backbone, where each \(\mathit{F}_{i}\) encodes view-specific features that are implicitly aligned within a shared canonical coordinate system.

\begin{figure}[t]  
    \centering  
    \includegraphics[width=\textwidth]{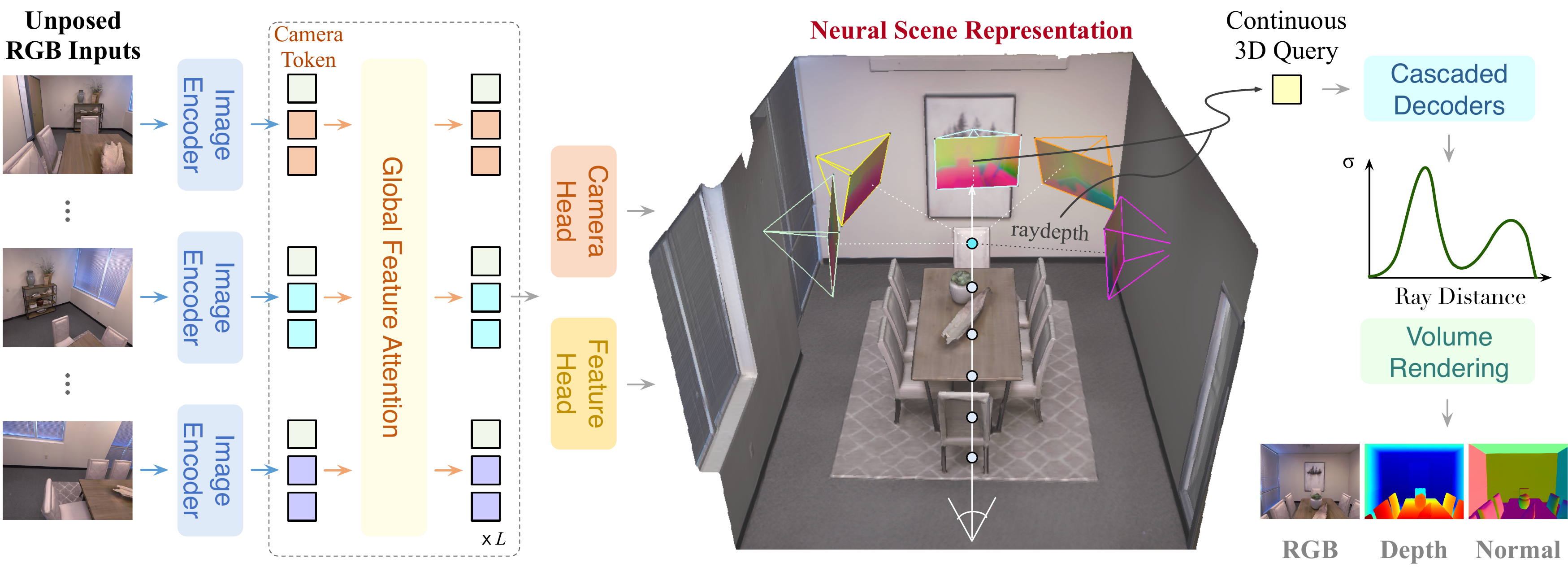}
    \vspace{-7mm}
    \caption{\textbf{Framework of IVGT.} Our model takes pose-free multi-view images as input and encodes them into per-view features, which are aggregated via global feature attention into a unified scene representation in a canonical coordinate system. This representation supports continuous 3D queries, where cascaded decoders predict SDF and colors for volume rendering and surface extraction.We additionally decode camera poses and per-view depth maps for the input views.}  
    \vspace{-5mm}
    \label{fig:main} 
\end{figure}

\textbf{Continuous 3D Query and Multi-view Aggregation.}
For an arbitrary 3D point \(\mathbf{x}\) defined in the canonical coordinate system of the first frame, we project it onto all input views using the corresponding camera parameters.
Suppose that \(\mathbf{x}\) is visible in \(N_k\) valid views, we first sample and aggregate the corresponding features from these views:
\begin{equation}
     \mathbf{z}_{f}=\sum_{i=1}^{N_k}\mathit{F}_{i}(\pi(\mathbf{x})).
\end{equation}
However, points located at different positions along the same projection ray may be mapped to the same pixel and thus share identical image features, while their underlying geometry and appearance can differ significantly.
Therefore, it is necessary to augment each queried point with positional information that reflects its spatial location.
A straightforward approach is to apply positional encoding directly on the 3D coordinates of \(\mathbf{x}\).
However, we argue that such encoding is ambiguous in our setting.
Specifically, the 3D coordinates of a certain point depend on the choice of the reference frame.
For the same scene, different choices of the reference frame would lead to different coordinate values for the same physical point, resulting in inconsistent positional encodings.
This may cause the network to rely on absolute coordinates and hinder its ability to learn consistent geometry.

Instead, we observe that what remains invariant is the relative position of a point with respect to each view, which can be represented by its ray depth.
We therefore encode the ray depths of \(\mathbf{x}\) with respect to all valid views using a lightweight MLP and aggregate them to obtain a ray-depth feature:
\begin{equation}
     \mathbf{z}_{d}=\sum_{i=1}^{N_k}f_\mathrm{raydepth}(d_{i}(\mathbf{x})),
\end{equation}
where \(d_{i}(\mathbf{x})\) is the ray depth of \(\mathbf{x}\) in each valid view and \(f_\mathrm{raydepth}\) denotes the MLP we design for this ray-depth embedding.
Finally, we concatenate the aggregated multi-view image features \(\mathbf{z}_{f}\) with the ray-depth feature \(\mathbf{z}_{d}\) to form the local spatial feature \(\mathbf{z}\) of \(\mathbf{x}\).

\textbf{Geometry and Appearance Decoding.}
Given the spatial feature of a queried point, we employ cascaded geometry and appearance decoders to predict its SDF value and color.
Specifically, we first use an 8-layer MLP to decode the SDF value \(\hat{s}\) and an intermediate appearance feature \(\hat{\mathbf{z}}\):
\begin{equation}
     (\hat{s},\hat{\mathbf{z}}) = f_{\theta }(\mathbf{z}).
\end{equation}
Then, we concatenate the appearance feature \(\hat{\mathbf{z}}\), the gradient \(\hat{\mathbf{n}}\) of the SDF value at this point, and a fixed positional encoding of the corresponding viewing direction \(\mathbf{v}\).
This concatenated feature is fed into a 2-layer MLP to predict the final color values:
\begin{equation}
     \hat{\mathbf{c}} = \mathbf{c}_{\theta }(\hat{\mathbf{z}}, \hat{\mathbf{n}}, \gamma(\mathbf{v})),
     \label{eq:color_prediction}
\end{equation}
where \(\gamma\) denotes the positional encoding following NeRF~\citep{mildenhall2021nerf}.

\subsection{Rendering and Surface Extraction}
\label{3.3}

\textbf{Volume Rendering.}
We optimize the neural scene representation described in Sec.~\ref{3.2} via a render-based reconstruction loss using differentiable volume rendering~\citep{oechsle2021unisurf, wang2021neus, yu2022monosdf}.
To render a pixel from input or novel viewpoints, we cast a ray \(\mathbf{r}\) from the camera center \(\mathbf{o}\) through the pixel along its view direction \(\mathbf{v}\). 
Along each ray, we sample \(M\) points \(\mathbf{x}_{\mathbf{r}}^{i}=\mathbf{o}+\mathit{t}_{\mathbf{r}}^{i}\mathbf{v}\) and predict their SDF values \(\hat{s}_{\mathbf{r}}^{i}\) and color values \(\hat{\mathbf{c}}_{\mathbf{r}}^{i}\).
Following VolSDF~\citep{yariv2021volume}, we transform the SDF values \(\hat{s}_{\mathbf{r}}^{i}\) to density values \({\sigma}_{\mathbf{r}}^{i}\) for volume rendering:
\begin{equation}
    \sigma_{\beta}(s)=\left\{\begin{array}{ll}
\frac{1}{2 \beta} \exp \left(\frac{s}{\beta}\right) & \text { if } s \leq 0 \\
\frac{1}{\beta}\left(1-\frac{1}{2} \exp \left(-\frac{s}{\beta}\right)\right) & \text { if } s>0
\end{array},\right.
\end{equation}
where \(\beta\) is a learnable parameter shared across different scenes.
Following NeRF~\citep{mildenhall2021nerf}, we compute the transmittance \(T_{\mathbf{r}}^{i}\) and alpha value \(\alpha_{\mathbf{r}}^{i}\) of sample point \(i\) along ray \(\mathbf{r}\):
\begin{equation}
    T_{\mathbf{r}}^{i}=\prod_{j=1}^{i-1}\left(1-\alpha_{\mathbf{r}}^{j}\right) \quad \alpha_{\mathbf{r}}^{i}=1-\exp \left(-\sigma_{\mathbf{r}}^{i} \delta_{\mathbf{r}}^{i}\right),
\end{equation}
where \(\delta_{\mathbf{r}}^{i}\) is the distance between neighboring sample points.
Then we can compute the color \(\hat{C}(\mathbf{r})\), the depth \(\hat{D}(\mathbf{r})\) and normal \(\hat{N}(\mathbf{r})\) of the surface intersecting the current ray as:
\begin{equation}
    \hat{C}(\mathbf{r})=\sum_{i=1}^{M} T_{\mathbf{r}}^{i} \alpha_{\mathbf{r}}^{i} \hat{\mathbf{c}}_{\mathbf{r}}^{i} \quad
    \hat{D}(\mathbf{r})=\sum_{i=1}^{M} T_{\mathbf{r}}^{i} \alpha_{\mathbf{r}}^{i} {t_{\mathbf{r}}^{i}} \quad
    \hat{N}(\mathbf{r})=\sum_{i=1}^{M} T_{\mathbf{r}}^{i} \alpha_{\mathbf{r}}^{i} \hat{\mathbf{n}}_{\mathbf{r}}^{i}.
\end{equation}

\textbf{Surface Extraction.}
During inference, we first project the depth maps predicted from the input-view features into the coordinate system of the first frame using the estimated camera poses, obtaining a coarse scene point cloud.
The spatial extent of the scene is then computed by the extrema of this point cloud.
Next, we uniformly sample a $64^3$ grid of points within this bounding volume and infer their SDF values.
Points whose absolute SDF values are below a predefined threshold are marked as valid.
We then upsample the grid to a higher resolution of $512^3$, retain only the valid regions, and evaluate their SDF values for surface reconstruction.
The final surface is extracted using the Marching Cubes algorithm~\citep{lorensen1998marching}.
Finally, we predict colors for the extracted surface points.
During color decoding for mesh vertices, the viewing direction required in Eq.~\ref{eq:color_prediction} is approximated using the surface normal direction at each surface point.

\subsection{Training}
\label{3.4}
\textbf{Training Losses.}
To stabilize the training of our implicit visual geometry transformer, we adopt a two-stage training strategy.
In the first stage, we supervise only the pose estimation and the rendered 2D outputs.
Specifically, we use the following loss functions:
\begin{equation}
\begin{aligned}
\mathcal{L}_{\mathrm{rgb}}
&= \sum_{\mathbf{r} \in \mathcal{R}} \|\hat{C}(\mathbf{r}) - C(\mathbf{r})\|_{1}, \\
\mathcal{L}_{\mathrm{depth-render}}
&= \sum_{\mathbf{r} \in \mathcal{R}} \|\hat{D}(\mathbf{r}) - D(\mathbf{r})\|_{1}, \\
\mathcal{L}_{\mathrm{normal}}
&= \sum_{\mathbf{r} \in \mathcal{R}} \|\hat{N}(\mathbf{r}) - \bar{N}(\mathbf{r})\|_{1}
+ \|1 - \hat{N}(\mathbf{r})^{\top} \bar{N}(\mathbf{r})\|_{1}, \\
\mathcal{L}_{\mathrm{camera}}
&= \sum_{i=1}^{N} \|\hat{\mathbf{g}}_{i} - \mathbf{g}_{i}\|_{\epsilon},
\end{aligned}
\label{eq:loss_stage1}
\end{equation}
where \(\mathcal{R}\) denotes the set of pixels in the sampled batch, \(C(\mathbf{r})\) is the observed pixel color, \(D(\mathbf{r})\) is the ground-truth depth normalized over the entire input sequence, and \(\bar{N}(\mathbf{r})\) is the predicted monocular normals transformed to the first-frame coordinate system. \(\mathbf{g}_{i}\) is the ground-truth camera parameters and \(\left | \cdot  \right |_{\epsilon}\) denotes the Huber loss.
The final loss used in the first stage is as follows:
\begin{equation}
    \mathcal{L}_{\mathrm{stage1}}=\mathcal{L}_{\mathrm{rgb}} + \lambda_1\mathcal{L}_{\mathrm{depth-render}}+\lambda_2\mathcal{L}_{\mathrm{normal}}+\lambda_3\mathcal{L}_{\mathrm{camera}}.
\end{equation}

In the second stage, we add an Eiknoal term~\citep{gropp2020implicit} and a smoothness term~\citep{oechsle2021unisurf, yu2022monosdf} to regularize SDF values to improve the stability of geometry reconstruction:
\begin{equation}
\begin{aligned}
\mathcal{L}_{\mathrm{eikonal}}&=\sum_{\mathbf{x} \in \mathcal{X}}\left(\left\|\nabla f_{\theta}(\mathbf{x})\right\|_{2}-1\right)^{2}, \\
\mathcal{L}_{\mathrm{smooth}} &= \sum_{\mathbf{x} \in \mathcal{X}} \left\| \frac{\nabla f_{\theta}(\mathbf{x})}{\|\nabla f_{\theta}(\mathbf{x})\|_2} - \frac{\nabla f_{\theta}(\mathbf{x}^{+})}{\|\nabla f_{\theta}(\mathbf{x}^{+})\|_2} \right\|_2,
\end{aligned}
\label{eq:loss_term}
\end{equation}
where \(\mathcal{X}\) are a set of uniformly sampled points together with near-surface points~\citep{yariv2021volume} and \(\mathbf{x}^{+}\) are neighboring points after adding small perturbations to the original sampled points.
And we also use the depth loss from VGGT~\citep{wang2025vggtvisualgeometrygrounded} to supervise the depth maps directly decoded from the input image features:
\begin{equation}
    \mathcal{L}_{\mathrm{depth}}=\sum_{i=1}^{N} \| \Sigma_{i}^{D} \odot \left(\hat{D}_{i}-D_{i}\right)\|+\| \Sigma_{i}^{D} \odot\left(\nabla \hat{D}_{i}-\nabla D_{i}\right) \|-\alpha \log \Sigma_{i}^{D},
\end{equation}
where \(\Sigma_{i}^{D}\) is the predicted uncertainty map between the predicted depth \(\hat{D}_{i}\) and the ground-truth depth \(D_{i}\), and \(\odot\) is the channel-broadcast element-wise product.
The final loss in the second stage is:
\begin{equation}
\mathcal{L}_{\mathrm{stage2}}=\mathcal{L}_{\mathrm{stage1}}+\lambda_4\mathcal{L}_{\mathrm{eikonal}}+\lambda_5\mathcal{L}_{\mathrm {smooth}}+\lambda_6\mathcal{L}_{\mathrm{depth}}.
\end{equation}

\textbf{Training Data.}
We train our IVGT using a diverse collection of datasets: ARKitScenes~\citep{baruch2021arkitscenes}, CO3Dv2~\citep{reizenstein2021common}, HyperSim~\citep{roberts2021hypersim}, MegaDepth~\citep{li2018megadepth}, OmniObject3D~\citep{wu2023omniobject3d}, ScanNet~\citep{dai2017scannet}, ScanNet++~\citep{yeshwanth2023scannet++}, Unreal4K~\citep{tosi2021smd} and WildRGBD~\citep{xia2024rgbd}.
These datasets cover objects and scenes, including both real and synthetic data.
We only require camera poses and RGB-D data as supervision (surface normals are obtained by predicting from RGB images using DSine~\citep{bae2024rethinking}), which allows us to fully leverage existing 3D datasets.

\textbf{Implementation Details.}
We initialize the image encoder and global feature extraction module with pre-trained weights from VGGT~\citep{wang2025vggtvisualgeometrygrounded}. 
We use the AdamW optimizer~\citep{loshchilov2017decoupled} and the learning rate warms up to a maximum value of 2e-4 and decreases according to a cosine schedule.
We set \(\lambda_1\), \(\lambda_2\), \(\lambda_3\), \(\lambda_4\), \(\lambda_5\), \(\lambda_6\) to 0.1, 0.05, 1.0, 0.01, 0.01, 0.1, respectively.
We render from 8 viewpoints (4 context and 4 novel) per iteration and sample 1024 rays per view.
We apply the error-bounded sampling strategy~\citep{yariv2021volume} to sample points along each ray.
We train our model on 4 A800 NVIDIA GPUs for 4 days.

%% file: chapters/4_exp.tex
\begin{table}[!t]
    \centering
    \caption{\textbf{Scene-level mesh reconstruction on ScanNet.} We compare against per-scene optimization baselines on Accuracy, Completeness, Chamfer Distance, Precision, Recall, and F-score. 
    }
    \vspace{-3mm}
    \footnotesize
    \resizebox{\textwidth}{!}{%
    \begin{tabular}{lccccccc}
    \toprule
    {} & Setting & Acc$\downarrow$ & Comp$\downarrow$ & Chamfer $\downarrow$ & Prec$\uparrow$ & Recall$\uparrow$ & F-score$\uparrow$\\
    \midrule
    COLMAP~\citep{colmapmvs}    & Per-Scene      & \underline{0.047} & 0.235 & 0.141 & \underline{0.711} & 0.441 & 0.537\\
    UNISURF~\citep{oechsle2021unisurf}         & Per-Scene      & 0.554 & 0.164 & 0.359 & 0.212 & 0.362 & 0.267\\
    NeuS~\citep{wang2021neus}                  & Per-Scene      & 0.179 & 0.208 & 0.194 & 0.313 & 0.275 & 0.291\\
    VolSDF~\citep{yariv2021volume}             & Per-Scene      & 0.414 & 0.120 & 0.267 & 0.321 & 0.394 & 0.346\\ 
    Manhattan-SDF~\citep{guo2022neural}        & Per-Scene      & 0.072 & 0.068 & 0.070 & 0.621 & 0.586 & 0.602\\
    MonoSDF~\citep{yu2022monosdf}              & Per-Scene      & \textbf{0.035} & \textbf{0.048} & \textbf{0.042} & \textbf{0.799} & \textbf{0.681} & \textbf{0.733}\\
    \textbf{IVGT}                                      & Generalizable & 0.069 & \underline{0.051} & \underline{0.060} & 0.663 & \underline{0.639} & \underline{0.647}\\
    \bottomrule
    \end{tabular}}
    \label{tab:scannet_mesh}
    \vspace{-5mm}
\end{table}

\begin{figure}[t]  
    \centering  
    \includegraphics[width=\textwidth]{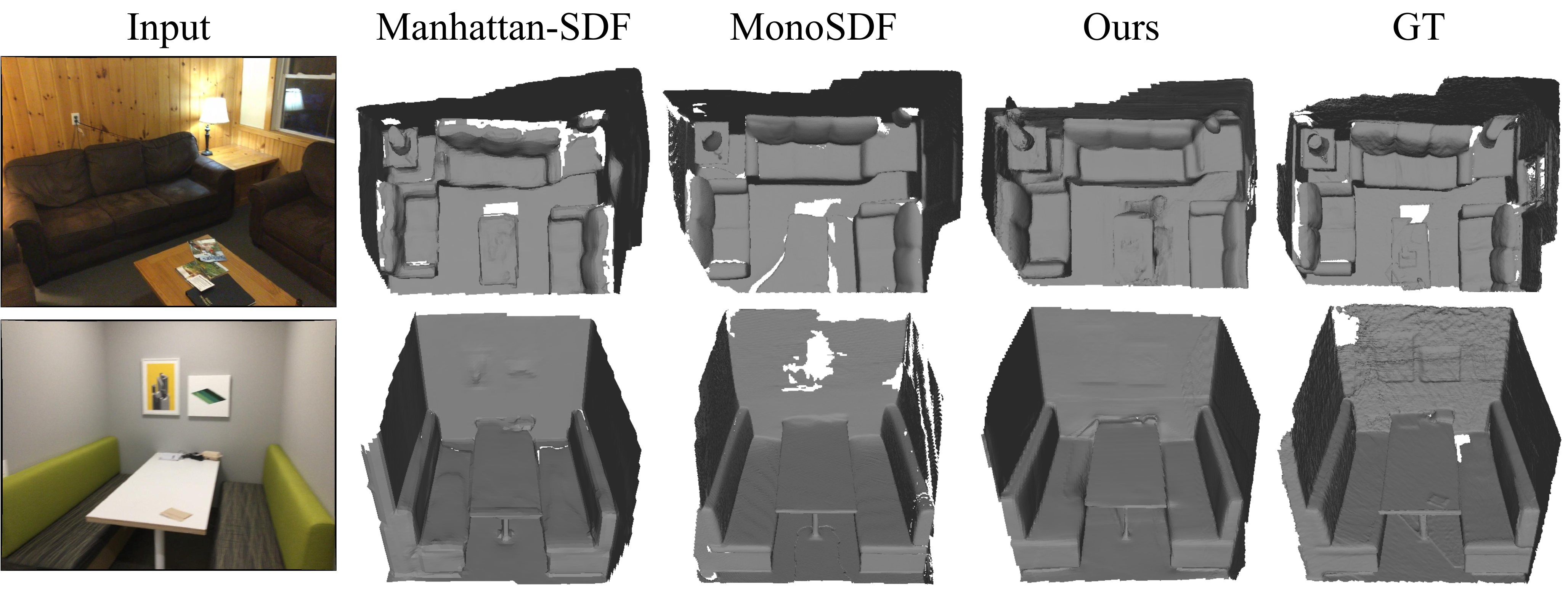}
    \vspace{-7mm}
    \caption{\textbf{Qualitative mesh reconstruction on ScanNet.} IVGT produces geometrically complete and surface-coherent meshes in a single forward pass, achieving comparable or superior reconstruction quality to per-scene optimization baselines.}
    \label{fig:meshmono} 
    \vspace{-5mm}
\end{figure}

\begin{figure}[t]  
    \centering  
    \vspace{-0mm}
    \includegraphics[width=\textwidth]{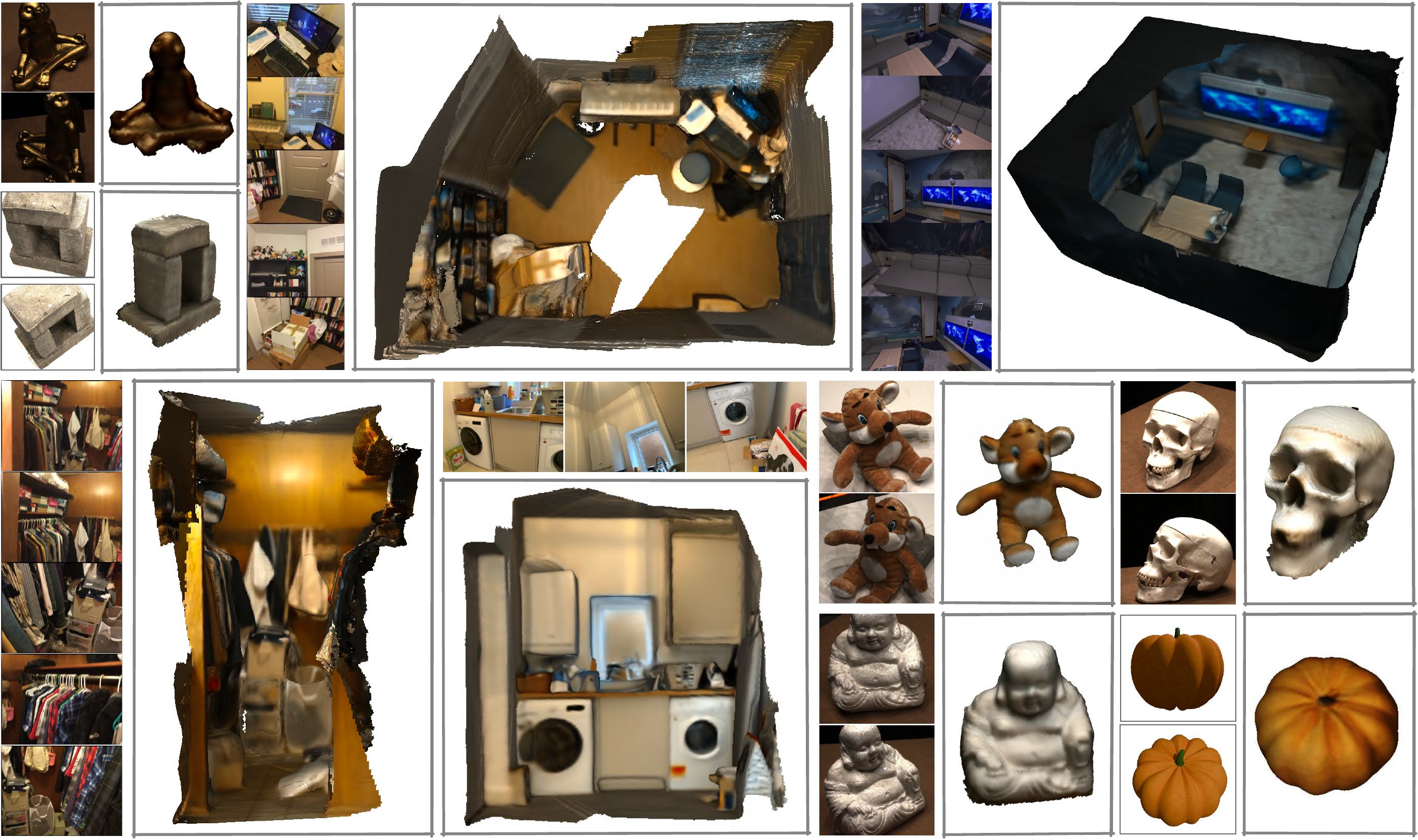}
    \vspace{-7mm}
    \caption{\textbf{Colored mesh reconstruction on diverse scenes and objects.} IVGT generalizes across indoor scenes and objects of varying scale, producing geometrically complete and visually consistent colored meshes without any test-time optimization.}
    \label{fig:meshcolor} 
    \vspace{-5mm}
\end{figure}

\begin{table*}[t]
\centering
\caption{\textbf{Pointmap reconstruction on 7-Scenes, NRGBD, and DTU.} We report mean and median Accuracy and Completeness (in meters for scene-level, mm for object-level). Pointmaps decoded directly from per-view features (\textbf{IVGT}) consistently outperform prior methods, while pointmaps projected from rendered depth maps (\textbf{IVGT} (from render)) also achieve competitive results.
}
\vspace{-3mm}
\small
\setlength{\tabcolsep}{3pt}
\resizebox{\textwidth}{!}{%
\begin{tabular}{lcccccccccccc}
\toprule
& \multicolumn{4}{c}{\textbf{7-Scenes} (scene, kf=50)}
& \multicolumn{4}{c}{\textbf{NRGBD} (scene, kf=100)}
& \multicolumn{4}{c}{\textbf{DTU} (object, kf=5)} \\
\cmidrule(lr){2-5} \cmidrule(lr){6-9} \cmidrule(lr){10-13}
Method
& \multicolumn{2}{c}{Acc. $\downarrow$}
& \multicolumn{2}{c}{Comp. $\downarrow$}
& \multicolumn{2}{c}{Acc. $\downarrow$}
& \multicolumn{2}{c}{Comp. $\downarrow$}
& \multicolumn{2}{c}{Acc. $\downarrow$}
& \multicolumn{2}{c}{Comp. $\downarrow$} \\
\cmidrule(lr){2-3} \cmidrule(lr){4-5}
\cmidrule(lr){6-7} \cmidrule(lr){8-9}
\cmidrule(lr){10-11} \cmidrule(lr){12-13}
& Mean & Med. & Mean & Med.
& Mean & Med. & Mean & Med.
& Mean & Med. & Mean & Med. \\
\midrule
Fast3R~\citep{yang2025fast3r}
& 0.053 & 0.023 & 0.084 & 0.033
& 0.080 & 0.039 & 0.075 & 0.039
& 3.605 & 1.855 & 3.048 & 1.274 \\
CUT3R~\citep{cut3r}
& 0.024 & 0.011 & 0.028 & \textbf{0.009}
& 0.086 & 0.038 & 0.047 & 0.016
& 4.749 & 2.588 & 2.764 & 1.176 \\
Point3R~\citep{wu2025point3r}
& 0.030& 0.015&  \underline{0.026}& \underline{0.010}& 0.065& 0.035& 0.030& 0.013& 6.111& 3.930& 2.880& 1.399\\
StreamVGGT~\citep{zhuo2025streaming}
& 0.046& 0.025& 0.032& 0.014& 0.080& 0.056& 0.044& 0.021& 2.520& 1.395& \underline{2.048}& \underline{1.075}\\
VGGT~\citep{wang2025vggtvisualgeometrygrounded}
& \underline{0.020} & \underline{0.008} & 0.028 & 0.012
& \underline{0.018} & \textbf{0.010} & \underline{0.017} & \underline{0.009}
& \textbf{1.305} & \textbf{0.743} & 2.435 & 1.737 \\
\textbf{IVGT}
& \textbf{0.016} & \textbf{0.007} & \textbf{0.021} & \textbf{0.009}
& \textbf{0.017} & \underline{0.011} & \textbf{0.015} & \textbf{0.006}
& \underline{1.686} & \underline{0.889} & \textbf{1.458} & \textbf{0.792} \\
\textbf{IVGT} (from render)
& 0.023 & 0.010 & 0.038 & 0.018
& 0.043 & 0.023 & 0.031 & 0.014
& 2.476 & 1.167 & 2.979 & 1.576 \\
\bottomrule
\end{tabular}%
}
\label{tab:pointmap_recon}
\vspace{-4mm}
\end{table*}

\begin{figure}[t]
    \centering  
    \includegraphics[width=\textwidth]{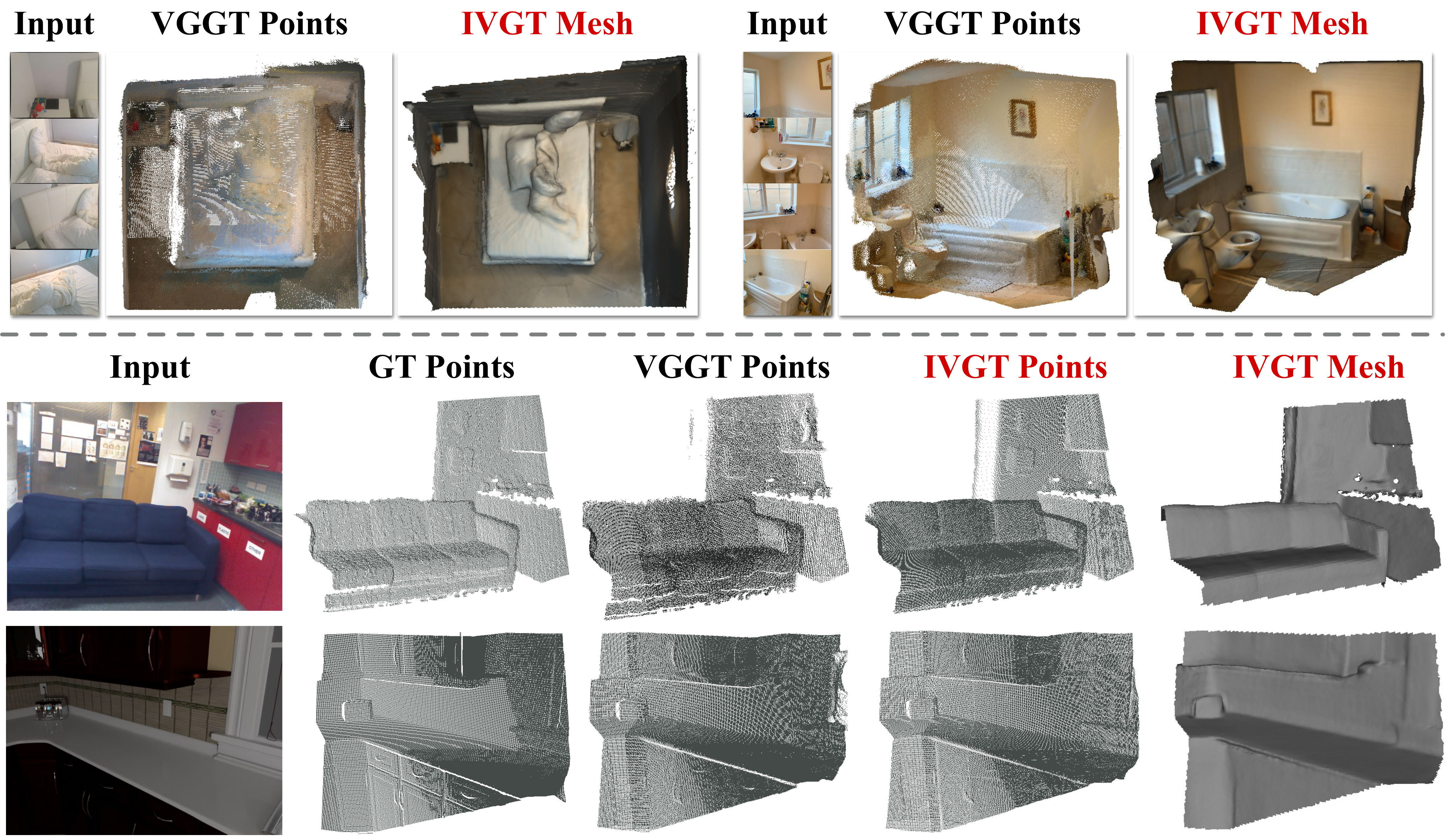}
    \vspace{-7mm}
    \caption{\textbf{Mesh vs.\ pointmap representations.} Pixel-aligned pointmap reconstruction suffers from sparsity and surface discontinuities, especially at object boundaries. Meshes and points extracted from IVGT exhibit significantly improved geometric continuity and completeness, validating the advantage of continuous implicit geometry over discrete explicit representations.}
    \label{fig:meshpoint}
    \vspace{-6mm}
\end{figure}

\section{Experiments}

We use various 2D/3D tasks (3D reconstruction, novel view synthesis, depth estimation, surface normal estimation, and camera pose estimation) to evalute our IVGT.
For each task, we select some representative methods as baselines and conduct comprehensive comparisons with our IVGT. 
We also provide various visualizations to demonstrate the effectiveness of our method.

\subsection{3D Mesh Reconstruction}
\label{4.1}
We compare against various baselines to show the performance of IVGT for mesh reconstruction.
On \textbf{ScanNet}~\citep{dai2017scannet}, we use the test split following Manhattan-SDF~\citep{guo2022neural} and MonoSDF~\citep{yu2022monosdf}.
In \Cref{tab:scannet_mesh}, we report the average reconstruction results for IVGT and baselines on several test scenes.
As a generalizable method, our method surpasses most per-scene optimization methods and ranks second only to MonoSDF. 
Given that MonoSDF requires several hours of optimization for each scene, IVGT offers a superior trade-off between computational efficiency and reconstruction quality. 
Qualitative comparisons with several representative baselines are shown in \Cref{fig:meshmono}.
We also visualize our mesh reconstruction results on a wider range of scene-level and object-level examples, as shown in \Cref{fig:meshcolor}.

\subsection{3D Pointmap Reconstruction}
IVGT estimates the relative poses of input images with respect to the first-frame coordinate system and predicts depth maps from the corresponding image features.
These depth maps are then used to reconstruct point maps of the input views within the first-frame coordinate system.
In addition, we can render depth maps for the input viewpoints from the learned neural scene representation.
These rendered depth maps similarly allow us to obtain pointmap reconstructions under the same coordinate system.
We evaluate the quality of pointmap reconstruction on both scene-level and object-level datasets, including \textbf{7-Scenes}~\citep{shotton2013scene}, \textbf{NRGBD}~\citep{azinovic2022neural}, and \textbf{DTU}~\citep{dtudataset} in \Cref{tab:pointmap_recon}.
Note that we only provide comparisons with a few representative methods for reference, as pointmap reconstruction is not the primary focus of IVGT but rather a by-product of our representation.
We can observe that the point maps reconstructed from depth maps directly decoded from image features achieve better performance than previous methods. 
The point maps obtained by projecting rendered depth maps also achieve results comparable to these representative methods.

 To further compare mesh and pointmap representations, we provide qualitative comparisons in \Cref{fig:meshpoint}. 
 Pixel-aligned pointmap reconstruction often suffers from sparsity and lack of surface continuity. 
 In contrast, meshes extracted from our implicit representation exhibit significantly improved surface coherence and geometric completeness.
 This demonstrates that IVGT enables not only accurate point-level reconstruction but also high-quality continuous surface modeling.
 
 \begin{table}[t] \footnotesize
\centering
\caption{\textbf{Camera pose estimation on ScanNet, Sintel, and TUM-dynamics.} We report Absolute Translation Error (ATE), Relative Translation Error (RPE trans), and Relative Rotation Error (RPE rot) after Sim(3) Umeyama alignment. IVGT achieves competitive or superior performance to state-of-the-art feed-forward methods across all three benchmarks.}
\vspace{-3mm}
\setlength{\tabcolsep}{2pt}
\resizebox{\textwidth}{!}{
\begin{tabular}{lccccccccc}
\toprule
\multirow{2}{*}{Method} 
& \multicolumn{3}{c}{ScanNet (Static)} 
& \multicolumn{3}{c}{Sintel} 
& \multicolumn{3}{c}{TUM-dynamics} \\
\cmidrule(r){2-4} \cmidrule(r){5-7} \cmidrule(r){8-10}
& ATE$\downarrow$ & RPE trans$\downarrow$ & RPE rot$\downarrow$
& ATE$\downarrow$ & RPE trans$\downarrow$ & RPE rot$\downarrow$
& ATE$\downarrow$ & RPE trans$\downarrow$ & RPE rot$\downarrow$ \\
\midrule
Fast3R~\citep{yang2025fast3r}
& 0.084& 0.059& 2.844& 0.274& 0.224& 15.69& 0.048& 0.052& 1.197\\
CUT3R~\citep{cut3r}
& 0.096& 0.022& 0.733& 0.210& 0.071& 0.627& 0.045& 0.014& 0.441\\
FLARE~\citep{zhang2025flare}
& 0.069& 0.025& 0.991& 0.158& 0.082& 2.863& \underline{0.026}& \underline{0.013}& 0.475\\
VGGT~\citep{wang2025vggtvisualgeometrygrounded}
& \underline{0.035}& \underline{0.015}& \textbf{0.403}& 0.169& 0.064&\textbf{0.478}& \textbf{0.012}& \textbf{0.010}& \underline{0.313}\\
WorldMirror~\citep{liu2025worldmirror}
& 0.037& 0.017& \underline{0.422}& \textbf{0.121}& \underline{0.063}& 0.535& \textbf{0.012}& \textbf{0.010}& \textbf{0.301}\\
\textbf{IVGT}
& \textbf{0.032}& \textbf{0.014}& \underline{0.422}& \underline{0.140}& \textbf{0.060}& \underline{0.530}& \textbf{0.012}& \textbf{0.010}& 0.319\\
\bottomrule
\end{tabular}
}
\label{tab:pose_estimation}
\vspace{-4mm}
\end{table}

\begin{table*}[t]
\centering
\caption{\textbf{Novel view synthesis on RealEstate10K and DL3DV.} We evaluate under sparse-view (2/8 views) and dense-view (32/64 views) settings and report PSNR, SSIM, and LPIPS. IVGT achieves competitive PSNR and SSIM scores, demonstrating that the learned implicit representation supports high-fidelity rendering from novel viewpoints.}
\vspace{-3mm}
\setlength{\tabcolsep}{1pt}
\resizebox{\textwidth}{!}{
\begin{tabular}{lcccccccccccc}
\toprule
\multirow{2}{*}{Method} 
& \multicolumn{3}{c}{RealEstate10K (2 views)} 
& \multicolumn{3}{c}{DL3DV (8 views)} 
& \multicolumn{3}{c}{RealEstate10K (32 views)} 
& \multicolumn{3}{c}{DL3DV (64 views)} \\
\cmidrule(lr){2-4} \cmidrule(lr){5-7} \cmidrule(lr){8-10} \cmidrule(lr){11-13}
& PSNR$\uparrow$ & SSIM$\uparrow$ & LPIPS$\downarrow$
& PSNR$\uparrow$ & SSIM$\uparrow$ & LPIPS$\downarrow$
& PSNR$\uparrow$ & SSIM$\uparrow$ & LPIPS$\downarrow$
& PSNR$\uparrow$ & SSIM$\uparrow$ & LPIPS$\downarrow$ \\
\midrule
FLARE~\citep{zhang2025flare} 
& 16.33 & 0.574 & 0.410 
& 15.35 & 0.516 & 0.591 
& -- & -- & -- 
& -- & -- & -- \\
AnySplat~\citep{jiang2025anysplat} 
& 17.62 & 0.616 & \underline{0.242} 
& 18.31 & 0.569 & \underline{0.258} 
& 19.96 & 0.718 & \underline{0.234} 
& 18.40 & 0.602 & \underline{0.286} \\
WorldMirror~\citep{liu2025worldmirror}
& \textbf{20.62} & \textbf{0.706} & \textbf{0.187} 
& \textbf{20.92} & \textbf{0.667} & \textbf{0.203} 
& \textbf{25.14} & \textbf{0.859} & \textbf{0.109} 
& \textbf{21.25} & \textbf{0.703} & \textbf{0.223} \\
\textbf{IVGT} 
& \underline{18.97} & \underline{0.656} & 0.449 
& \underline{19.74} & \underline{0.627} & 0.494 
& \underline{21.91} & \underline{0.751} & 0.394 
& \underline{19.62} & \underline{0.659} & 0.515 \\
\bottomrule
\end{tabular}
}
\label{tab:nvs_re10k_dl3dv}
\vspace{-5mm}
\end{table*}

\begin{figure}[t!]  
    \centering  
    \includegraphics[width=\textwidth]{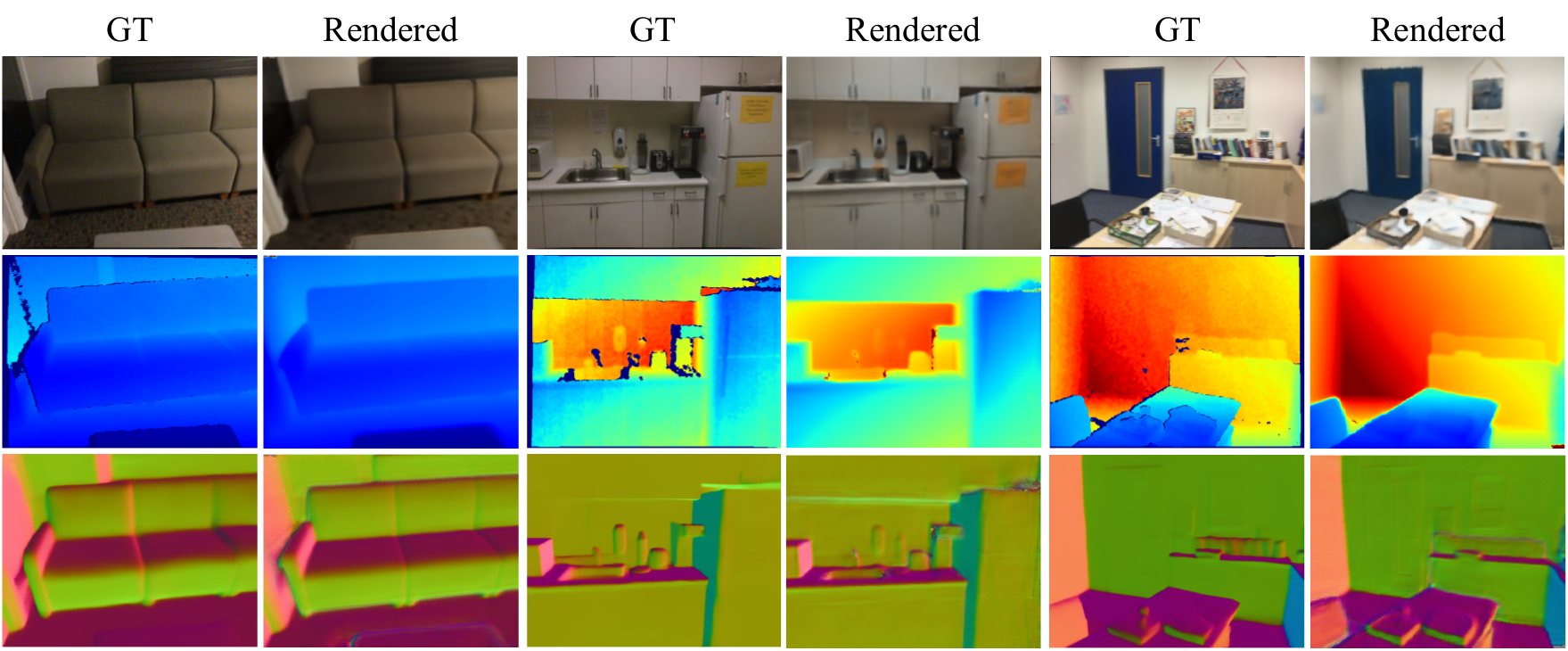}
    \vspace{-7mm}
    \caption{\textbf{Novel view synthesis on ScanNet.} Given pose-free multi-view inputs, IVGT renders RGB images, depth maps, and surface normal maps from unseen viewpoints. The outputs are visually coherent and geometrically smooth, confirming that the unified implicit representation captures both appearance and 3D structure. All three modalities are derived from the same SDF field without any task-specific decoding heads, demonstrating the versatility of our continuous implicit geometry.}
    \label{fig:2d_render} 
    \vspace{-5mm}
\end{figure}

\subsection{Camera Pose Estimation}
Following MonST3R~\citep{zhang2024monst3r} and CUT3R~\citep{cut3r}, we evaluate the camera pose estimation performance on \textbf{ScanNet}, \textbf{Sintel} and \textbf{TUM-dynamics}, as shown in \Cref{tab:pose_estimation}.
We report Absolute Translation
Error (ATE), Relative Translation Error (RPE trans), and Relative Rotation Error (RPE rot) after
applying a Sim(3) Umeyama alignment with the ground truth.
Our method achieves comparable or even superior performance to representative feed-forward approaches on camera pose estimation metrics across multiple datasets.

\subsection{Novel View Synthesis}
We use \Cref{tab:nvs_re10k_dl3dv} to report the quantitative evaluation results for novel view synthesis under the feed-forward setting.
For \textbf{RealEstate10K}~\citep{zhou2018stereo}, we use 200 sampled test scenes following NopoSplat~\citep{ye2024no} and WorldMirror~\citep{liu2025worldmirror}.
We use 3 novel views per scene in the sparse-view setting and 4 novel views per scene in the dense-view setting.
For \textbf{DL3DV}~\citep{ling2024dl3dv}, we follow the FLARE~\citep{zhang2025flare} test split and evaluate in 112 unseen scenes.
Each test scene contains 9 novel views in both sparse-view and dense-view settings.
Our method achieves performance comparable to existing feed-forward methods on this task. 

We visualize the novel view synthesis results on the \textbf{ScanNet} test scenes in \Cref{fig:2d_render}. 
Given multi-view input images, IVGT renders RGB images, depth maps, and surface normal maps from unseen viewpoints using the learned scene representation. 
The rendered results are visually consistent with the ground truth, while depth maps and surface normal maps exhibit smooth and coherent structures, which indicate that the representation captures both geometry and appearance in a unified manner.

\subsection{Monocular and Video Depth Estimation}
We compare with several representative feed-forward methods in \Cref{tab:depth_nyu_sintel}.
We evaluate the monocular depth estimation on \textbf{NYUv2}~\citep{silberman2012indoor} and \textbf{Sintel}~\citep{butler2012naturalistic} datasets, and evaluate the video depth estimation on \textbf{Sintel} dataset.
The depth maps directly decoded from multi-view features achieve better performance in depth estimation metrics than prior methods, while the depth maps rendered from the neural scene representation obtain comparable results under the same metrics, demonstrating that the implicit representation preserves accurate geometry.

\begin{table}[t]
\centering
\caption{\textbf{Monocular and video depth estimation on NYUv2 and Sintel.} We report Abs Rel and $\delta\!<\!1.25$ for both monocular and video settings. Depth maps decoded directly from per-view features (\textbf{IVGT}) outperform prior methods, while depth maps rendered from the implicit field (\textbf{IVGT} (from render)) remain competitive, demonstrating complementary strengths of the two decoding pathways.}
\vspace{-3mm}
\resizebox{\textwidth}{!}{
\begin{tabular}{lcccccc}
\toprule
\multirow{2}{*}{Method} 
& \multicolumn{2}{c}{NYUv2 (Monocular)} 
& \multicolumn{2}{c}{Sintel (Monocular)} 
& \multicolumn{2}{c}{Sintel (Video)} \\
\cmidrule(r){2-3} \cmidrule(r){4-5} \cmidrule(r){6-7}
& Abs Rel$\downarrow$ & $\delta < 1.25 \uparrow$
& Abs Rel$\downarrow$ & $\delta < 1.25 \uparrow$
& Abs Rel$\downarrow$ & $\delta < 1.25 \uparrow$ \\
\midrule
Fast3R~\citep{yang2025fast3r} & 0.093 & 0.898 & 0.544 & 0.509 & 0.638 & 0.422 \\
CUT3R~\citep{cut3r} & 0.081 & 0.914 & 0.418 & 0.520 & 0.417 & 0.507 \\
FLARE~\citep{zhang2025flare} & 0.089 & 0.898 & 0.606 & 0.402 & 0.729 & 0.336 \\
VGGT~\citep{wang2025vggtvisualgeometrygrounded} & \textbf{0.056} & \underline{0.951} & 0.606 & \underline{0.599} & \underline{0.299} & \underline{0.638} \\
\textbf{IVGT} & \underline{0.063} & \textbf{0.954} & \textbf{0.309} & \textbf{0.620} & \textbf{0.295} & \textbf{0.646} \\
\textbf{IVGT} (from render) & 0.067 & 0.945 & \underline{0.398} & 0.570 & 0.542 & 0.582 \\
\bottomrule
\end{tabular}}
\label{tab:depth_nyu_sintel}
\vspace{-4mm}
\end{table}

\begin{table}[t]
\centering
\caption{\textbf{Surface normal estimation on NYUv2 and iBims-1.} We report mean/median angular error and the percentage of pixels within $22.5^\circ$ and $30^\circ$ angular thresholds. IVGT achieves results comparable to dedicated normal estimation methods, demonstrating that the SDF-based representation captures high-quality surface geometry as a natural by-product of multi-view reconstruction.}
\vspace{-3mm}
\resizebox{\textwidth}{!}{
\begin{tabular}{lcccccccc}
\toprule
\multirow{2}{*}{Method} 
& \multicolumn{4}{c}{NYUv2} 
& \multicolumn{4}{c}{iBims-1} \\
\cmidrule(r){2-5} \cmidrule(r){6-9}
& mean$\downarrow$ & med$\downarrow$ & $22.5^\circ \uparrow$ & $30^\circ \uparrow$
& mean$\downarrow$ & med$\downarrow$ & $22.5^\circ \uparrow$ & $30^\circ \uparrow$ \\
\midrule
OASIS~\citep{chen2020oasis} & 29.2 & 23.4 & 48.4 & 60.7 & 32.6 & 24.6 & 46.6 & 57.4 \\
EESNU~\citep{bae2021estimating} & \textbf{16.2} & \underline{8.5} & 77.2 & \underline{83.5} & 20.0 & 8.4 & 73.4 & 78.2 \\
Omnidata v2~\citep{kar20223d} & 17.2 & 9.7 & 76.5 & 83.0 & \underline{18.2} & \underline{7.0} & \underline{77.4} & \underline{81.1} \\
DSine~\citep{bae2024rethinking} & \underline{16.4} & \textbf{8.4} & \textbf{77.7} & \underline{83.5} & \textbf{17.1} & \textbf{6.1} & \textbf{79.0} & \textbf{82.3} \\
\textbf{IVGT} & 16.6 & 10.4 & \underline{77.3} & \textbf{84.2} & 20.1 & 10.6 & 74.4 & 79.6 \\
\bottomrule
\end{tabular}}
\vspace{-3mm}
\label{tab:normal}
\end{table}

\subsection{Surface Normal Estimation}
Following DSine~\citep{bae2024rethinking}, we evaluate surface normal estimation on \textbf{NYUv2} and \textbf{iBims-1}, as shown in \Cref{tab:normal}.
We report the angular error between predicted and ground-truth normal maps, including mean and median errors, as well as the percentage of pixels with angular error below $22.5^\circ$ and $30^\circ$.
We compare our method with representative normal estimation approaches, such as OASIS~\citep{chen2020oasis}, EESNU~\citep{bae2021estimating}, Omnidata v2~\citep{kar20223d} and DSine~\citep{bae2024rethinking}. 
Our results are comparable to these strong baselines, demonstrating that our model can recover consistent surface geometry despite being primarily optimized for multi-view 3D reconstruction rather than dedicated normal estimation.

\subsection{Ablation and More Analysis}

\textbf{Ablation on the Positional Information Injecting Choices.}
As discussed in Sec.~\ref{3.2}, directly encoding positional information using absolute 3D coordinates may introduce ambiguity due to the dependence on the choice of reference frame. 
To further validate this, we conduct an ablation study on ScanNet, where we train four variants with different positional information injecting strategies under exactly the same training setup. 
Each setting is trained for 16 hours.
\textit{No Embed} means using no spatial encoding, \textit{XYZ-PosEnc} refers to applying fixed positional encoding on 3D coordinates, and \textit{XYZ-Embed} means directly embedding coordinates using an MLP.
We evalute the mesh reconstruction performance of these four variants on several test scenes we use in Sec.~\ref{4.1} and report the average metrics in \Cref{tab:scannet_ablation}. 
The \textit{No Embed} variant fails to learn meaningful geometry, as most points are predicted with positive SDF values, resulting in empty space and no valid mesh for evaluation, highlighting the necessity of positional information. 
Both \textit{XYZ-PosEnc} and \textit{XYZ-Embed} variants rely on absolute coordinates and achieve limited performance, likely due to ambiguity introduced by reference-frame-dependent representations. 
In contrast, our \textit{Raydepth-Embed} design achieves the best performance across all metrics by encoding view-consistent relative geometry.

\begin{figure}[t!]  
    \centering  
    \includegraphics[width=\textwidth]{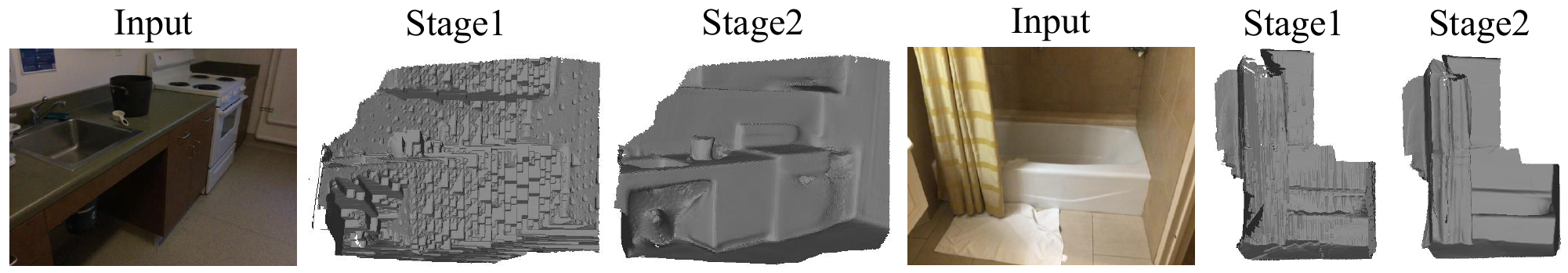}
    \vspace{-7mm}
    \caption{\textbf{Effect of the two-stage training strategy.} The first stage (2D supervision only) captures coarse geometry but produces rough, irregular surfaces. Adding Eikonal and smoothness regularization in the second stage significantly improves surface quality, yielding smooth and coherent meshes.
     }
    \label{fig:eik_ablation} 
    \vspace{-5mm}
\end{figure}

\textbf{Analysis on the Training Setup.}
We adopt a two-stage training strategy to balance geometric learning and surface regularization. 
Directly introducing 3D constraints in the early stage tends to hinder the model from learning the underlying geometric distribution, while relying solely on 2D supervision leads to rough and irregular surfaces. 
We therefore first train the model using only 2D supervision to capture coarse geometry, and then introduce Eikonal and smoothness regularization in the second stage to enforce surface consistency. 
As shown in \Cref{fig:eik_ablation}, the first stage results exhibit noticeable surface irregularities, whereas the second stage produces significantly smoother and more coherent surfaces, validating the effectiveness of our training setup.

\begin{table}[!t]  \footnotesize
    \centering
    \caption{\textbf{Ablation on positional encoding strategies for 3D query points.} We compare XYZ-PosEnc (fixed encoding on absolute 3D coordinates), XYZ-Embed (MLP on 3D coordinates), and our Raydepth-Embed (MLP on view-relative ray depths). Absolute-coordinate encodings introduce reference-frame ambiguity and degrade performance, while our view-consistent ray depth encoding achieves the best results across all metrics.}
    \vspace{-3mm}
    \setlength{\tabcolsep}{12pt}
    \begin{tabular}{lcccccc}
    \toprule
    {} & Acc$\downarrow$ & Comp$\downarrow$ & Chamfer $\downarrow$ & Prec$\uparrow$ & Recall$\uparrow$ & F-score$\uparrow$\\
    \midrule
    XYZ-PosEnc         
    & 0.624 & 0.158 & 0.391 & 0.220 & 0.346 & 0.260 \\
    XYZ-Embed                  
    & 0.494 & 0.148 & 0.321 & 0.272 & 0.364 & 0.306 \\
    Raydepth-Embed             
    & \textbf{0.219} & \textbf{0.103} & \textbf{0.161} & \textbf{0.434} & \textbf{0.431} & \textbf{0.429} \\ 
    \bottomrule
    \end{tabular}
    \label{tab:scannet_ablation}
    \vspace{-4mm}
\end{table}

%% file: chapters/5_conclusion.tex
\section{Conclusion and Discussions }
In this paper, we have presented IVGT, an implicit visual geometry transformer that learns a pose-free neural scene representation and enables continuous spatial queries for joint geometry and appearance modeling. 
Unlike prior explicit models that decode discrete, pixel-aligned pointmaps, IVGT models the scene as a global implicit SDF field that supports continuous 3D queries at arbitrary locations.
Extensive experiments demonstrate that our method achieves strong performance across multiple tasks, including mesh reconstruction, point cloud reconstruction, novel view synthesis, depth and surface normal estimation, and camera pose estimation, while generalizing across diverse scenes. 
Our formulation bridges feed-forward multi-view representation learning with neural implicit geometry modeling, representing a step toward scalable and pose-free continuous surface reconstruction.

\textbf{Limitations.}
Despite the promising results, IVGT has several limitations for future exploration:
\begin{itemize}[
    topsep=0pt,     %
    partopsep=0pt,  %
    itemsep=0pt,    %
    leftmargin=1em, %
]
	\item While our SDF-based rendering produces geometrically coherent outputs, the rendering quality in terms of high-frequency appearance details lags behind dedicated novel view synthesis methods such as those based on 3D Gaussian splatting~\citep{liu2025worldmirror}, as reflected by the quantitative results. 
	The smoothness prior enforced by Eikonal regularization, while beneficial for surface coherence, suppresses fine-grained geometric structures such as thin objects and sharp edges.
	\item The current framework assumes a static scene and a bounded spatial extent, limiting its applicability to dynamic environments or large-scale unbounded outdoor scenes.
	\item The continuous 3D query mechanism requires projecting each sampled point onto all input views and aggregating features, which is more computationally intensive than direct pixel-aligned decoding at inference time, which may hinder real-time deployment.
\end{itemize}
Addressing these (e.g., incorporating appearance-oriented decoding heads, handling dynamic content, or developing more efficient query strategies) remains an important direction for future research.